\documentclass{article}
\usepackage{ijcai22}

\usepackage{times}

\usepackage{soul}
\usepackage{url}
\usepackage[hidelinks]{hyperref}
\usepackage[utf8]{inputenc}
\usepackage[small]{caption}
\usepackage{graphicx}
\usepackage{amsmath}
\usepackage{booktabs}
\usepackage[ruled,vlined]{algorithm2e}
\usepackage{tikz}
\usepackage{subcaption}
\usepackage{array}
\usepackage{setspace}
\newcolumntype{H}{>{\setbox0=\hbox\bgroup}c<{\egroup}@{}}
\title{Machine Learning Methods in Solving the Boolean Satisfiability Problem}

\author{
	Wenxuan Guo$^1$\and
	Junchi Yan$^1$\footnote{Correspondence author.}\and
	Hui-Ling Zhen$^2$\and
	Xijun Li$^2$\and
	Mingxuan Yuan$^2$\And
	Yaohui Jin$^{1*}$\\
	\affiliations
	$^1$Department of CSE, and MoE Key Lab of Artificial Intelligence, Shanghai Jiao Tong University\\
	$^2$Noah's Ark Lab, Huawei Ltd.\\
	\emails 
	{
		\{arya\_g,yanjunchi,jinyh\}@sjtu.edu.cn, \{xijun.li,zhenhuiling2,yuan.mingxuan\}@huawei.com
}}

\begin{document}
	\begin{spacing}{0.95}

	\maketitle
	
	\begin{abstract}
		This paper reviews the recent literature on solving the Boolean satisfiability problem (SAT), an archetypal NP-complete problem, with the help of machine learning techniques. Despite the great success of modern SAT solvers to solve large industrial instances, the design of handcrafted heuristics is time-consuming and empirical. Under the circumstances, the flexible and expressive machine learning methods provide a proper alternative to solve this long-standing problem. We examine the evolving ML-SAT solvers from naive classifiers with handcrafted features to the emerging end-to-end SAT solvers such as NeuroSAT, as well as recent progress on combinations of existing CDCL and local search solvers with machine learning methods. Overall, solving SAT with machine learning is a promising yet challenging research topic. We conclude the limitations of current works and suggest possible future directions.
	\end{abstract}
	
	\section{Introduction}
	The Boolean satisfiability problem, often referred to as SAT, is the first proven NP-complete problem~\cite{cook1971complexity} in the field of computational complexity. This hard combinatorial problem consistently attracts researchers' attention for its wide application and that a variety of problems can be reduced to SAT. For theoretical interests, numerous combinatorial problems can be expressed in propositional formulae and solved by running a SAT solver~\cite{iwama1994sat}, e.g. graph coloring~\cite{velev2007exploiting}, vertex cover~\cite{plachetta2021sat} and clique detection~\cite{skansi2020clique}. It also serves as a useful tool for automated theorem proving, one typical case of which is the resolution of Keller’s conjecture~\cite{brakensiek2020resolution}.
	Moreover, there are plenty of industrial applications of SAT solving, such as bounded model checking, configuration management, and equivalence checking in circuit design. Hence,  SAT solving not only promotes research progress but also enables more economical workflow.
	
	\begin{figure}[tb!]
	    \centering
	    \includegraphics[width=\columnwidth]{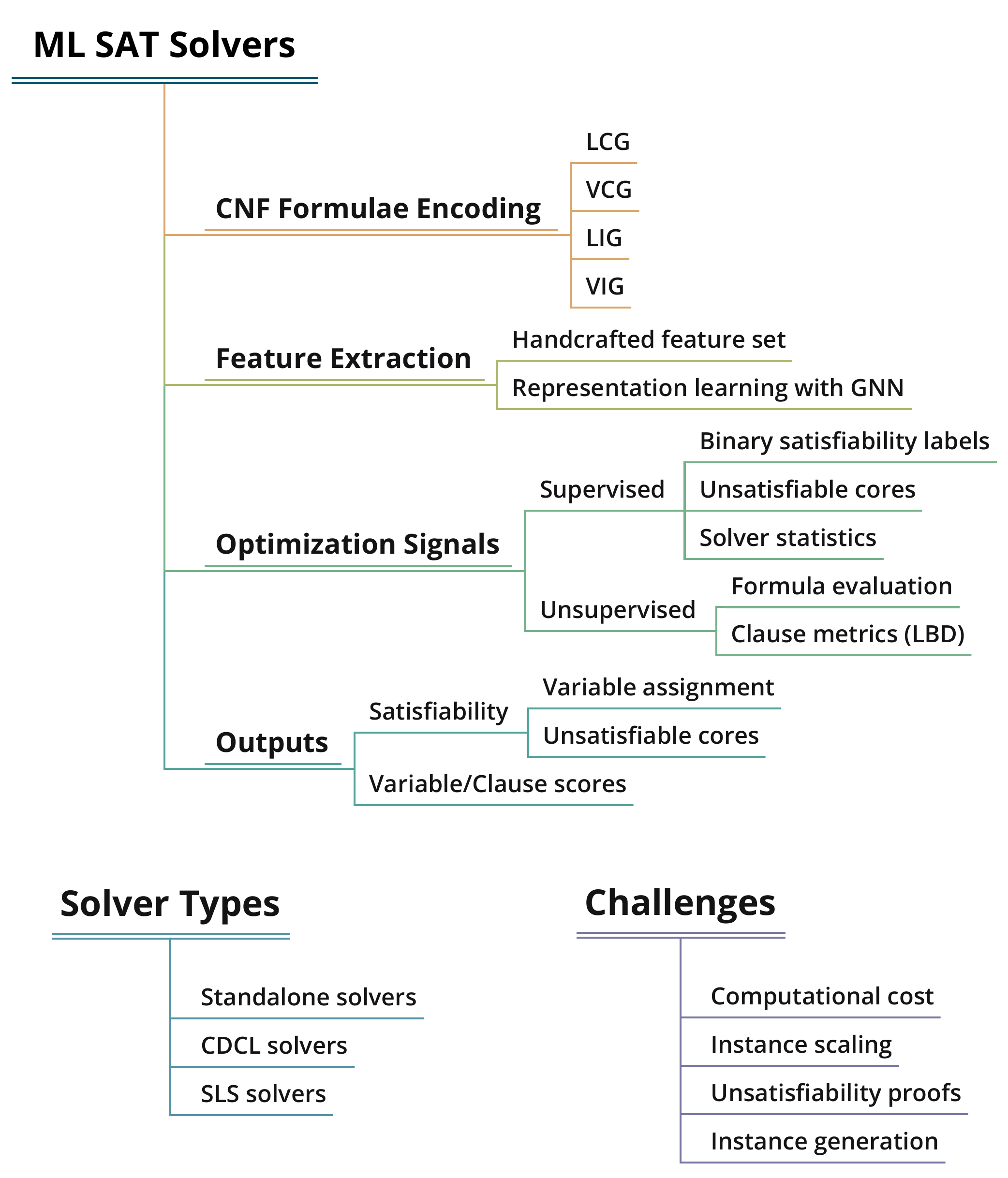}
	    \caption{Overview of SAT solvers with ML techniques.}
	    \label{fig:overview}
	\end{figure}
	
	Since the P versus NP problem remains unsettled, researchers respect the difficulty of the SAT problem and struggle to design efficient SAT solvers. Modern solvers mostly follow the paradigms proposed by~\cite{davis1962machine} and~\cite{marques1999grasp} in the last century, and groundbreaking SAT solving patterns and performance leaps are absent in recent years. Practitioners have been focusing on incremental heuristics based on existing solver paradigms, which requires a comprehensive understanding of SAT solvers. 
	Meanwhile, machine learning (ML), especially the surging deep learning techniques have advanced into the combinatorial optimization field and yielded a number of promising new avenues of research~\cite{bengioMachineLearningCombinatorial2020}.
	In this case, it is natural for the SAT community to seek an integration of machine learning and SAT solving, which enables automatic and organic deduction and saves human labor.
	
	Currently, there are primarily three patterns for this combination to boost SAT solving: 1) standalone SAT solvers with pure ML methods; 2) replacing some components of existing CDCL solvers with learning-directed heuristics, and 3) modifying the local search solvers with learning-aided modules. 
	Moreover, ML techniques help remove the limitation that SAT instances are mainly from SAT competition. SAT instance generation aims to provide sufficient representative training/test samples in industrial scenarios. Different from the generation of random and combinatorial instances, generating pseudo-industrial instances requires following certain structural characteristics, as well as the problem scale. Typical examples of pseudo-industrial SAT instance generation include SATGEN~\cite{wuLearningGenerateIndustrial2019} and G2SAT~\cite{youG2SATLearningGenerate2019}, where SATGEN uses an unsupervised generative model that implicitly portrays the intrinsic features, and G2SAT proposes a node-merging (-splitting) algorithm to generate bipartite graphs from (to) a forest.
    
	We also note a few related surveys on SAT~\cite{SATSurvey17,SatSurvey21}, while the former is learning-free, and the latter covers a broader scope in terms of symbolic logic. To our best knowledge, there lacks a focused survey covering the topic of solving SAT with machine learning.
	This survey encompasses directly optimizing SAT solving with the aid of machine learning techniques, e.g. MLP, naive Bayes, and neural networks, in the aforementioned three ways. Portfolio solvers and algorithm runtime prediction are not discussed in this paper, as it is a general technique applicable to other problems as well (see~\cite{hutterAlgorithmRuntimePrediction2014} for a survey). The extensions of SAT, e.g. maximum satisfiability problem (MAX-SAT), satisfiability modulo theories (SMT), and quantified Boolean formula problem (QBF), are also beyond the scope of this survey.
	
	\section{Preliminaries}
	This section starts with basic definitions of the SAT problem, followed by some classic learning-free SAT solvers which serve as paradigms for creating new solvers. It will highlight the key heuristics suitable for ML-based modification.
	
	\subsection{Boolean Satisfiability Problem}
	In propositional logic, a Boolean formula is built from Boolean variables (only allowed to take value $\mathsf{True}$ or $\mathsf{False}$) and three logic operators: conjunction ($\wedge$), disjunction ($\vee$) and negation ($\neg$). The Boolean satisfiability problem aims to determine whether there exists a way of variable assignment so that a given Boolean formula evaluates to $\mathsf{True}$. In the positive case, the formula is \textit{satisfiable}, as opposed to \textit{unsatisfiable} ones. If a SAT instance is satisfiable, it only takes polynomial time to verify an assignment. Otherwise, the formula may contain an \textit{unsatisfiable core}, a subset of clauses whose conjunction is still unsatisfiable.
	
	Since every propositional formula can be transformed into an equivalent formula in conjunctive normal form (CNF), we only consider this form in the following discussion. A formula in CNF consists of a conjunction of \textit{clauses}, where each clause is a disjunction of \textit{literals}, a \textit{variable} or its negation. 
	
	The complexity of SAT has been studied by the Cook–Levin theorem~\cite{cook1971complexity}, stating that SAT is NP-complete. In other words, if there exists a deterministic polynomial algorithm, then every NP problem can be solved by a deterministic polynomial algorithm. Currently, SAT problems are mostly solved by optimized searching-based methods, with exponential worst-case complexity.
	
	\subsection{Mainstream Classic SAT Solvers}
	Formally, a \textit{solver} is a procedure aiming to solve the SAT problem: given an input of Boolean formula, a solver is supposed to yield the judgment of its satisfiability and provide a valid assignment if it is satisfiable. A \textit{complete} solver is able to deduce that a SAT instance is unsatisfiable with a proof, as opposed to \textit{incomplete} algorithms (see~\cite{heule2015proofs} for more details on unsatisfiability proofs).
	
	\begin{algorithm}[tb!]
		\SetKwProg{Fn}{Function}{}{end}
		\SetKwFunction{cdcl}{CDCL}
		\SetKwFunction{unp}{UnitPropagation}
		\SetKwFunction{ava}{AllVariablesAssigned}
		\SetKwFunction{ca}{ConflictAnalysis}
		\SetKwFunction{pbv}{PickBranchingVariable}
		\SetKwFunction{bt}{Backtrack}
		\SetKw{conflict}{CONFLICT}
		\SetKwInOut{input}{Input}
		\SetKwIF{If}{ElseIf}{Else}{if}{}{else if}{else}{end if}%
		
		\input{A CNF formula $\varphi$; partial assignment $\nu$}
		\Fn{\cdcl($\varphi$, $\nu$)} {
			\lIf{\unp($\varphi$, $\nu$) == \conflict}{\KwRet UNSAT}
			$dl\leftarrow 0$\;
			\While{not \ava($\varphi$, $\nu$)} {
				($x$, $v$) = \pbv($\varphi$, $\nu$)\;
				$dl\leftarrow dl + 1$\;
				$\nu\leftarrow\nu\cup\{x,v\}$\;
				\If{\unp($\varphi$, $\nu$) == \conflict} {
					$\beta=$ \ca($\varphi$, $\nu$)\;
					\lIf{$\beta<0$}{\KwRet UNSAT} 
					\Else{
						\bt($\varphi$, $\nu$, $\beta$)\;
						$dl\leftarrow\beta$\;
					}
				}
			}
		}
		\caption{Typical CDCL algorithm$^\dagger$}
		\label{alg:cdcl}
		\small $^\dagger$ adapted from~\cite{biereHandbookSatisfiability2009}
	\end{algorithm}
	
	\subsubsection{Conflict-Driven Clause Learning (CDCL) Solvers}
	The conflict-driven clause learning (CDCL) algorithm, first proposed in the solver GRASP~\cite{marques1999grasp}, is a representative complete SAT algorithm. As an improvement of the Davis–Putnam–Logemann–Loveland (DPLL) algorithm~\cite{davis1962machine}, its backbone is a backtracking-based search algorithm that selects a variable at a time for tentative assignment and backtracks chronologically once the reduced formula contains an empty clause. The primary heuristic of the CDCL algorithm, as suggested by its name, is that it learns new clauses from conflicts (invalid partial assignments which lead to unsatisfiability) and add them to the original clause. The standard organization of a CDCL SAT solver is described in Algorithm~\ref{alg:cdcl}, to which we pay special attention to the following two concepts.
	
	\paragraph{Variable Selection}
	Variable selection heuristics, or branching heuristics, find the most ``valuable'' unassigned variable to branch on. Popular candidates include the variable state independent decaying sum (VSIDS) heuristic~\cite{moskewicz2001chaff} and its variants, where a score is recorded for each variable, and each time the variable with the greatest score is selected. When a clause is learned by a CDCL solver, the score of involved variables is increased by some amount. At regular intervals, a procedure called \textit{rescoring} is executed. For example, all scores are divided by some constant. In this way, variables in more conflicts recently are preferred.
		
	\begin{algorithm}[tb!]
		\SetKwProg{Fn}{Function}{}{end}
		\SetKwFunction{wst}{WalkSAT}
		\SetKwInOut{input}{Input}
		\SetKwInOut{param}{Parameters}
		\SetKwIF{If}{ElseIf}{Else}{if}{}{else if}{else}{end if}%
		
		\input{A CNF formula $\varphi$}
		\param{Intergers $MaxTries, MaxFlips$; noise parameter $p\in[0,1]$}
		\Fn{\wst($\varphi$)} {
			\For{$i\leftarrow 1$ \KwTo $MaxTries$}{
				$\sigma\leftarrow$ a random truth assignment for $\varphi$\;
				\For{$j\leftarrow 1$ \KwTo $MaxFlips$}{
					\lIf{$\sigma$ satisfies $\varphi$} {\KwRet $\sigma$}
					$C\leftarrow$ a random unsatisfied clause of $\varphi$\;
					\uIf{$\exists$ variable $x\in C$ with break-count $=0$} {
						$v\leftarrow x$\;
					}
					\Else{
						With probability $p$: \\
						\quad $v\leftarrow $ a random variable in $C$\;
						With probability $1-p$: \\
						\quad $v\leftarrow $ a variable in $C$ with the smallest break-count\;
					}
					Flip $v$ in $\sigma$\;
				}
			}
			\KwRet FAIL\;
		}
		\caption{WalkSAT~\protect\cite{selman1993local}$^\ddag$}
		\label{alg:walksat}
		\small $^\ddag$ adapted from~\cite{biereHandbookSatisfiability2009}
	\end{algorithm}
	
	\paragraph{Literal Block Distance and Glue Clauses}
	CDCL solvers benefit from learning from conflicts, but this convenience costs more memory consumption: the number of learned clauses grows exponentially, and therefore it must perform clause deletion regularly.
	Literal block distance (LBD) is a metric proposed by~\cite{audemard2009predicting}, defined as the number of distinct decision levels of the variables in a clause. LBD can measure the quality of clauses, due to the empirical observation that decision levels regularly decrease during search~\cite{audemard2009predicting}. It also points out that clauses with an LBD of 2 are of vital importance, thus termed as ``glue clauses''.
	
%

	\subsubsection{Stochastic Local Search (SLS) Solvers}
	Stochastic local search (SLS) algorithms are effective for solving random and hard combinatorial instances, a typical example of which is WalkSAT~\cite{selman1993local}, as shown in Algorithm~\ref{alg:walksat}. As an incomplete solver, it starts from an initial variable assignment and flips the value of a selected variable at each iteration, until a legal assignment is found or the time limit is exceeded. To avoid getting trapped in the local minima, stochastic restarts are performed during the search if the restart criterion is met. The key heuristics involved in an SLS solver is the restart policy, initialization scheme, and variable selection for flipping. For example, GSAT~\cite{selman92anew} does not restart and chooses the variable that minimizes the number of unsatisfied clauses after flipping. Sec.~\ref{sec:SLS} discusses how to create new heuristics for SLS solvers with machine learning techniques. 
	
	\subsection{Graph Representation of Boolean Formulae}
	\label{sec:graph_rep}
\begin{figure}[tb!]
		\centering
		\begin{subfigure}[c]{0.24\linewidth}
			\parbox[][2.5cm][c]{\linewidth}{
				\centering
				\includegraphics[width=0.9\columnwidth]{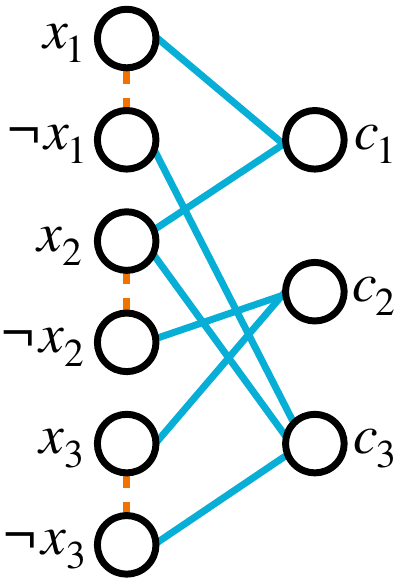}
			}
			\subcaption{LCG}
		\end{subfigure}
		\begin{subfigure}[c]{0.22\linewidth}
			\parbox[][2.5cm][c]{\linewidth}{
				\centering
				\includegraphics[width=0.9\columnwidth]{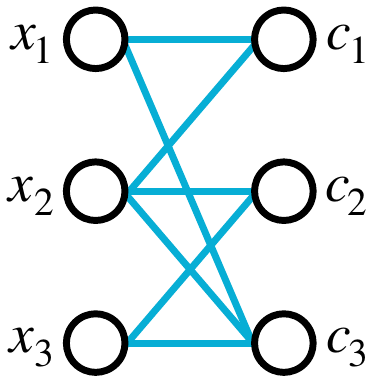}
			}
			\subcaption{VCG}
		\end{subfigure}
		\begin{subfigure}[c]{0.28\linewidth}
			\parbox[][2.5cm][c]{\linewidth}{
				\centering
				\includegraphics[width=0.9\columnwidth]{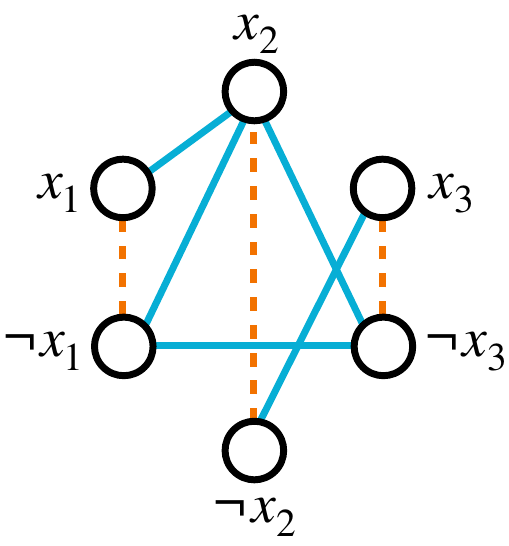}
			}
			\subcaption{LIG}
		\end{subfigure}
		\begin{subfigure}[c]{0.22\linewidth}
			\parbox[][2.5cm][c]{\linewidth}{
				\centering
				\includegraphics[width=0.9\columnwidth]{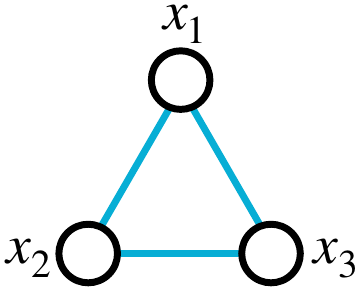}
			}
			\subcaption{VIG}
		\end{subfigure}
		\caption{Four graph representations of the propositional formula $(x_1\vee x_2)\wedge(\neg x_2\vee x_3)\wedge(\neg x_1\vee x_2 \vee \neg x_3)$. The dashed lines denote the common connections between complementary literals in GNN for message passing.}
		\label{fig:cnf_graph}
	\end{figure}

	To apply graph neural networks to the SAT problem, the first step is to encode CNF formulae into graphs. There are four straightforward graph representations of a CNF formula: 1) literal-clause graph (LCG), 2) literal-incidence graph (LIG), 3) variable-clause graph (VCG) and 4) variable-incidence graph (VIG). LCG is a bipartite graph with literals on one side and clauses on the other, with edges connecting literals to the clauses where they occur, while LIG consists only of literal nodes and two literals have an edge if they co-occur in a clause. VCG and VIG are defined similarly by merging the positive and negative literals of the same variables. An illustration of four graph representations is shown in Fig.~\ref{fig:cnf_graph}. The decreasing complexity of the four graphs suggests an increasing level of information compression: one can recover the original CNF formula from an LCG without loss, but barely characterize the formula given a VIG. Therefore, LCG and LIG are preferred in practice.
	
	\section{Towards Machine Learning of SAT Solving}
	In this section, we first discuss standalone SAT solvers and then review the ML components in CDCL and SLS solvers.
	
	\subsection{Standalone SAT Solvers}
	\label{sec:standalone}
	If we treat the SAT problem as a classification task, many machine learning models can serve as the classifier as long as we first extract features from input formulae, which has been tried over a decade ago. Deep learning, on the other hand, changes the way of feature extraction and facilitates end-to-end frameworks to predict satisfiability.
	
	\subsubsection{Classifiers with Handcrafted Features}
	\label{sec:classifier}
	The successful portfolio SAT solver SATzilla~\cite{xuSATzillaPortfoliobasedAlgorithm2008} constructed a 48-dimensional feature set and used ridge regression to fit a runtime prediction function for further algorithm selection. This feature set was manually designed to describe the property of an instance, from basic information like problem size to variable graph features. Despite the limitations of human intervention, initial statistical methods were inspired to utilize this feature set along with basic machine learning models (e.g. MLP, decision tree, naive Bayes, etc.)~\cite{devlinSatisfiabilityClassificationProblem2008,danisovszkyClassificationSATProblem2020} to classify SAT instances into binary categories for satisfiability prediction. \cite{devlinSatisfiabilityClassificationProblem2008} trained and evaluated a variety of machine learning models on crafted, industrial, and random instances from SAT competitions and SATLib, achieving accuracy above 90\% on most benchmarks. \cite{danisovszkyClassificationSATProblem2020} built another 48-dimensional feature set with an emphasis on special problems and clause properties. The authors experimented on different structures of neural networks as well as basic machine learning classifiers and achieved the best result of about 99\% accuracy.
	
	Although this line of work achieved high accuracy on different benchmarks, a major drawback of this method occurs in the features extraction part. The SATzilla-style feature set includes DPLL and local-search probing features, and thus an instance is tried to be solved by different solvers before the classification. Moreover, the time for feature extraction of~\cite{devlinSatisfiabilityClassificationProblem2008} can be high as thousands of seconds for one instance, comparable to the runtime of a complete solving routine. Therefore, a more promising approach would be constructing a module that analyzes the input formula directly and independently.

	\subsubsection{End-to-End Neural SAT Solvers}
	A turning point of feature extraction for SAT formulae occurred with the emergence of deep learning, especially graph neural networks (GNNs), which lifts the limitation that input features are handcrafted and involve expert knowledge. The graph representations of CNF formulae and graph neural networks liberate human experts from feature engineering, allowing the GNN to extract useful embedding automatically, and enable end-to-end learning frameworks.
	
	\cite{bunzGraphNeuralNetworks2017} made an early attempt from the aspect of natural language processing: whether the CNF formulae can be treated as sentences in natural language with recursive neural networks (RNNs), but it led to failure. As another trial, it used LIG representation for GNN and one-hot edge features to differentiate clauses apart. Since the difficulty of random 3-SAT instances is sensitive to the clause-to-atom ratio, it was tested in three settings and achieved an accuracy of roughly 65\%, quite above a random baseline and indicating a promising direction.
	
	\begin{figure}[tb!]
		\centering
		\includegraphics[width=\columnwidth]{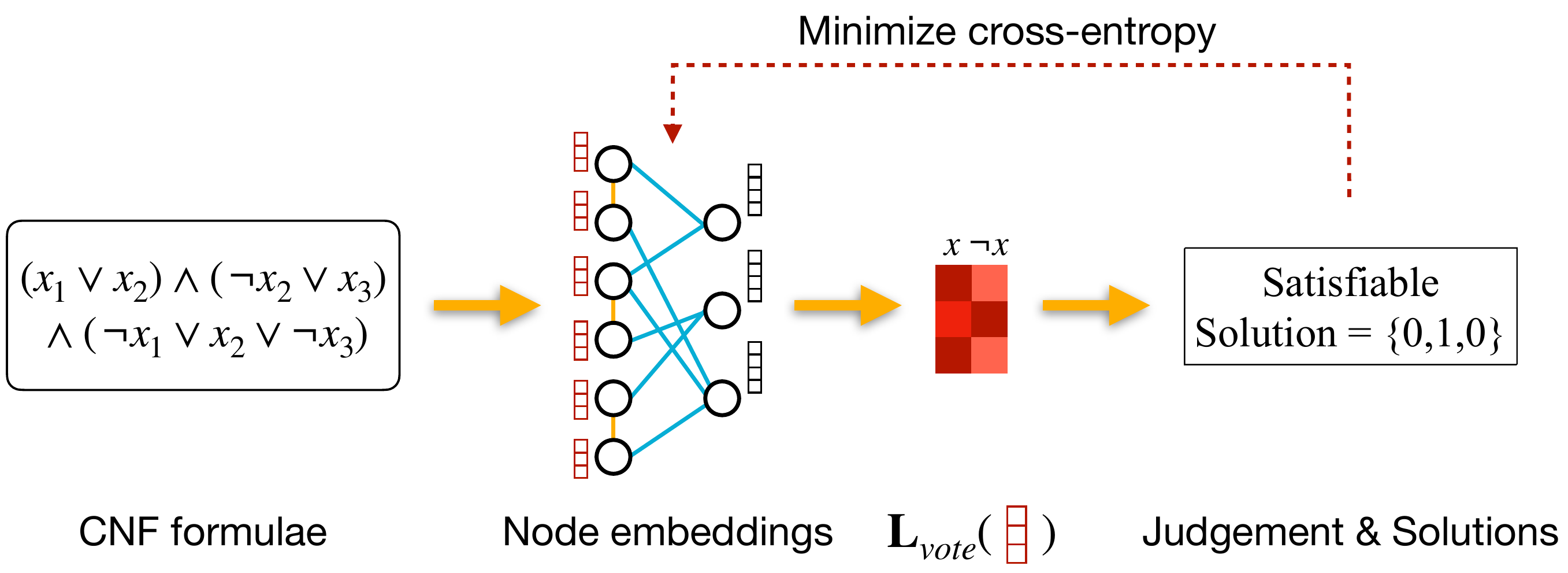}
		\caption{Pipeline of the end-to-end SAT solver NeuroSAT~\protect\cite{selsamLearningSATSolver2019}. Input CNF formulae are transformed into LCGs, and each literal and clause updates their embeddings from neighbors and complements. $\mathbf{L}_{vote}$ computes a vote scalar for each literal, which are aggregated to yield the final classification. Darker red of literals indicates a higher score in voting \textit{satisfiable}, and the solution chooses the more confident literal from each complementary pair.}
		\label{fig:neurosat}
	\end{figure}
	
	The seminal work NeuroSAT~\cite{selsamLearningSATSolver2019} (shown in Fig.~\ref{fig:neurosat}) improves the above results and presents an end-to-end framework to predict satisfiability on random instances by message passing neural network (MPNN). Different from sentences in natural language,  Boolean formulae have their unique properties of permutation invariance and negation invariance\footnote{The satisfiability of a formula is not affected by permuting the variables, the clauses or the literals within a clause. It is also not affected by negating every literal corresponding to a given variable.~\cite{selsamLearningSATSolver2019}}, which are preserved in NeuroSAT by symmetric edge connection and message passing. Specifically, the CNF formulae are encoded as LCGs, and node embedding is iteratively updated in a two-stage fashion for clauses and literals. First, each clause updates the embedding by receiving messages from neighboring literals. Next, each literal receives messages from neighboring clauses and the complementary literal. 
	At the final layer, a scalar \textit{vote} is computed for each literal that represents its confidence in predicting the formula to be \textit{satisfiable}, and the mean vote value decides the final output. 
	The training and test dataset of NeuroSAT is $\mathbf{SR}(n)$ ($10\le n\le 40$), which consists of pairs of random SAT problems on $n$ variables such that one element of the pair is satisfiable, the other is unsatisfiable, and they differ by negating one literal occurrence in a single clause. Although the networks are trained in a supervised way only with the label of satisfiability, NeuroSAT attempts to yield a solution for instances with positive prediction. On $\mathbf{SR}(40)$, NeuroSAT reached an accuracy of 85\% and solved 70\% of SAT problems.
	
	Despite NeuroSAT's impressive performance on random instances, its training paradigm bears limitations. First, it demands millions of training samples, which is inefficient and inconsistent with small instances it can solve (typically at most 40 variables). Second, satisfiability of instances is required beforehand and must be computed by other solvers, compromising the meaning of training a new solver for elementary instances. Hence, subsequent works prefer an unsupervised way and challenge more complex benchmarks.
	
	
	
	QuerySAT~\cite{ozolinsGoalAwareNeuralSAT2021} develops a recurrent neural SAT solver that is trained in an unsupervised fashion. By relaxing the variables to continuous values $[0,1]$, the unsupervised loss $\mathcal{L}_{\phi}(x)$ for a formula $\phi$ is defined as
    \begin{equation}
		V_{c}(x)=1-\prod_{i \in c^{+}}\left(1-x_{i}\right) \prod_{i \in c^{-}} x_{i}, \ 
		\mathcal{L}_{\phi}(x)=\prod_{c \in \phi} V_{c}(x), \label{eq:unsupervised_loss}
	\end{equation}
	where $x_i$ is the value of the $i$-th variable and $c^+$ gives the set of variables that occur in the clause $c$ in the positive form and $c^-$ in the negated form. The authors proved that this loss function is sufficient to uniquely identify the SAT formula $\phi$.
	Different from \cite{selsamLearningSATSolver2019}, this loss function is not only used at the final layer but also calculated for each query: at every time step, QuerySAT produces a query and evaluates a loss along with its gradient w.r.t. the query, which are then used for updating state vectors.
	The model is optimized towards minimizing the sum of all losses and produces a variable assignment. For empirical validation, QuerySAT used multiple benchmarks including $k$-SAT, 3-SAT, and some combinatorial problems and achieved accuracy over 90\%.
	
	In a similar vein, DG-DAGRNN~\cite{amizadehLearningSolveCircuitSAT2018} concentrates on the circuit satisfiability problem (Circuit-SAT), a special form of SAT, by unsupervised learning. This study proposed a neural Circuit-SAT solver which can harness structural information in the input circuits. 
	The framework consists of a neural functional $\mathcal{F}_\theta$ which contains an embedding function, an aggregation function, and a classification function. To implement a fully differentiable training strategy, they proposed an explore-exploit mechanism as in reinforcement learning. Specifically, they use the smooth min and max functions instead of hard ones in min-max circuits to allow the gradients to flow through all paths in the input circuit. Finally, they define a satisfiability function to check if the resulting assignment satisfies the circuit. 
	Following the settings in~\cite{selsamLearningSATSolver2019}, 
	DG-DAGRANN could converge much faster than NeuroSAT. When the number of variables is much larger, the performance of NeuroSAT declines faster than DG-DAGRNN as the number of variables increases. In graph $k$-coloring decision problem, DG-DAGRNN could solve 48\% and 27\% of the SAT problems in two generated datasets, respectively, while NeuroSAT failed to solve any of them, even when the number of iterations is big enough (128 propagation iterations).
	
	\begin{table*}[tb!]
		\centering
		\resizebox{\linewidth}{!}{
			\begin{tabular}{ccccc}
				\toprule
				\textbf{Methods} & \textbf{Networks} & \textbf{Learning} & \textbf{Solver Type} & \textbf{Instance Type} \\
				\midrule
				\cite{bunzGraphNeuralNetworks2017} & GNN & Supervised & Standalone & 3-SAT \\
				NeuroSAT~\cite{selsamLearningSATSolver2019} & GNN \& LSTM & Supervised & Standalone & $\mathbf{SR}(n)$ \\
				QuerySAT~\cite{ozolinsGoalAwareNeuralSAT2021} & GNN \& Recurrent & Unsupervised & Standalone & $k$-SAT \& Combinatorial \\
				DG-DAGRNN~\cite{amizadehLearningSolveCircuitSAT2018} & DG-DAGRANN & Unsupervised & Standalone & $k$-SAT \& Combinatorial\\
				NeuroCore~\cite{selsamGuidingHighperformanceSAT2019} & GNN & Supervised & CDCL & SATCOMP \\
				\cite{jaszczurNeuralHeuristicsSAT2020} & GNN \& Attention & Supervised & DPLL \& CDCL & $\mathbf{SR}(n)$ \\
				Graph-$Q$-SAT~\cite{kurinCanLearningGraph2020} & GNN & Reinforcement & CDCL & 3-SAT \\
				NeuroGlue~\cite{hanEnhancingSATSolvers2020} & GNN & Supervised \& Reinforcement & CDCL & SATCOMP \\
				GVE~\cite{zhangEliminationMechanismGlue2021} & GNN & Reinforcement & CDCL & SATCOMP \\
				NeuroCuber~\cite{hanLearningCubingHeuristics2020} & GNN & Supervised & Cube-and-conquer & Combinatorial\\
				NeuroComb~\cite{wangNeuroCombImprovingSAT2021} & GNN & Supervised & CDCL & SATCOMP \\
				\cite{yolcuLearningLocalSearch2019} & GNN & Reinforcement & SLS & 3-SAT \& Combinatorial \\
				NLocalSAT~\cite{zhangNLocalSATBoostingLocal2020} & GGCN & Supervised & SLS & Random \\
				\bottomrule
		\end{tabular}}
		\caption{Summary of reviewed SAT solvers built with machine learning techniques. }
		\vspace{-10pt}
		\label{tab:comparison}
	\end{table*}
	
	\subsection{Learning-aided CDCL Solvers}
	The full-stack SAT solvers with machine learning discussed in Sec.~\ref{sec:standalone} are more of methodological than practical interests: they are trained with millions of samples and tested only on small random or combinatorial benchmarks with no guarantee of correctness, and thus fail to function for industrial purposes. Therefore, a more incremental way is to modify existing CDCL solvers and replace the bottleneck components with machine learning modules. In practice, there are few suitable candidates for such a modification if we take into account the considerable computation time for neural networks.
	The most popular direction is the branching heuristics~\cite{selsamGuidingHighperformanceSAT2019,jaszczurNeuralHeuristicsSAT2020,kurinCanLearningGraph2020,hanEnhancingSATSolvers2020,zhangEliminationMechanismGlue2021,hanLearningCubingHeuristics2020,wangNeuroCombImprovingSAT2021}, plus some works on optimizing initialization~\cite{wuImprovingSATsolvingMachine2017}, clause deletion~\cite{vaezipoorLearningClauseDeletion2020} and restart policy~\cite{liangMachineLearningBasedRestart2018}.
	
	
	\subsubsection{Variable Initialization}
	In search-based CDCL algorithms, the variables branched on are assigned to binary value $\mathsf{True}$ or $\mathsf{False}$ based on some initialization scheme. The most basic initialization is by random. \cite{wuImprovingSATsolvingMachine2017} assumed that an initial value close to solutions can provide considerable speedup for solving the problem, and it proposed to train a logistic regression module to predict satisfiability of 3-SAT formulae with 10 predefined features as input. The preferred initialization value for each variable is determined by a series of Monte Carlo trials with satisfiability prediction from the trained predictor. The author reported a decrease of 23\% in runtime for satisfiable instances if preprocessing time is not counted, which even outweighs the decrease in runtime. The essence of this method is very similar to classifiers in Sec.~\ref{sec:classifier}, and the logistic regression predictor can be replaced by peer methods.
	
	\subsubsection{Branching Heuristics}
	After the standalone neural network SAT solver NeuroSAT, the following work, a more economical model NeuroCore~\cite{selsamGuidingHighperformanceSAT2019}, proposed to incorporate NeuroSAT into Minisat, a CDCL solver that implements the EVSIDS heuristic (a variant of VSIDS) and keeps an activity score for each variable. NeuroCore integrates the satisfiability prediction in NeuroSAT by periodically replacing the activity scores with the output from neural networks, termed as \textit{periodic refocusing}. NeuroCore's framework is made up of three MLPs, one for updating the clause embedding based on the literals in it, one for updating the literal embedding based on the clauses it is in, and one for computing the scores for each variable as the output of NeuroCore. Different from the original NeuroSAT, the networks are trained with a focus on the unsatisfiable core. The logic behind this is that variables in the unsatisfiable core are prone to lead to conflicts, and thus are valuable for branching. \cite{selsamGuidingHighperformanceSAT2019} generated a dataset mapping unsatisfiable problems to the variables which are in the unsatisfiable cores. The hybrid solver neuro-minisat solved 10\% more problems than Minisat on SATCOMP-2018 within the standard timeout of 5000 seconds, and a similar improvement was observed on Glucose.
	
	Other works inspired by the NeuroSAT framework combine the GNN module with CDCL solvers to determine the variable to branch on. 
	For example, \cite{jaszczurNeuralHeuristicsSAT2020} uses a similar network architecture as NeuroSAT and predicts satisfiability for each literal as well as the whole formula.
	Graph-$Q$-SAT~\cite{kurinCanLearningGraph2020} utilizes reinforcement learning instead of supervised learning for label efficiency. It formulates the Boolean formulae as VCGs and learns a value function for each variable node, with a straightforward policy to select the variables with max value.
	
	
	Besides direct supervision of satisfiability and crafted unsupervised loss, another approach is to use statistics produced by solvers as supervision, such as the LBD and glue variables (those that are likely to occur in glue clauses). 
	NeuroGlue~\cite{hanEnhancingSATSolvers2020} trains a neural network that predicts the glue variables. The authors follow NeuroCore~\cite{selsamGuidingHighperformanceSAT2019} and apply the periodic refocusing technique on the state-of-the-art SAT solver, CaDiCaL~\cite{biere2017cadical}, and replace the EVSIDS activity scores with network outputs. The training data is generated by running CaDiCaL and counting the numbers of times each variable appears in glue clauses, used as supervision for glue variable prediction. There is also a reinforcement learning module that selects variables sequentially in an episode. The reward favors small glue levels. 
	GVE~\cite{zhangEliminationMechanismGlue2021} uses two separate modules to determine branching variables and their values. There is a GNN-based glue variable selector by RL similar to~\cite{hanEnhancingSATSolvers2020} and another LSTM module that predicts the value of variables. Finally, the simplified CNF formulae are sent to a deterministic solver. Both NeuroGlue and GVE are tested on industrial benchmarks. NeuroGlue improves on the PAR-2 score of CaDiCaL, while the complex architecture of GVE increases running time significantly.
	
	Besides the conflict-driven pattern, there is also a variable selection heuristic in the cube-and-conquer paradigm~\cite{heule2017solving}. This technique aims to reduce the complexity of the SAT solver by partitioning a SAT problem into subproblems (cube), which are then solved (conquered) by CDCL solvers in parallel, and there is a variable selection heuristic for cubing. Each selection will add two new leaves to the search tree that correspond to different assignments of the variable. Then the cutoff heuristic is used to check the new formulas and freeze some leaves if they are easy for CDCL. NeuroCuber~\cite{hanLearningCubingHeuristics2020} applies the network architectures of NeuroCore to the cube-and-conquer framework with an emphasis on DRAT proof occurrence counts. Besides the variable scoring head and the clause scoring head in NeuroSAT, NeuroCuber has another variable scoring head that predicts occurrence counts of variables in DRAT proofs, which can be roughly thought of as a compressed representation of resolution trees. In \cite{hanLearningCubingHeuristics2020}, it is assumed that if a variable occurs frequently in a resolution tree, branching on it would minimize the average size of the resolution trees (and proportionally the solving times) for the leaves. Through experiments on datasets of unsatisfiable problems, they show that models trained to predict DRAT variable counts usually outperform those trained to predict the occurrence of a variable in an unsatisfiable core.
	
	Most above works fall short either of applicability to industrial problems or of computational efficiency. NeuroComb~\cite{wangNeuroCombImprovingSAT2021} proposes to embed GNN prediction into CDCL solvers in a more balanced way. To reduce the cost of periodic refocusing in NeuroCore, it adopts offline predictions computed before launching the CDCL solver. During the searching process, the dynamic branching heuristic (e.g. VSIDS) is periodically interrupted by this static information for a short time, so that the heuristic is under a constant but slow influence of GNN predictions.
	
	
	\subsubsection{Restart Policy}
	Restarts are not only useful for SLS solvers but also effective for CDCL solvers. During the search process, a restart occurs when a certain number of conflicts are met and the solver discards the current partial assignment but keeps the learned clauses and search from the start again. \cite{liangMachineLearningBasedRestart2018} designs a new restart policy called machine learning-based restart (MLR) that triggers a restart when the predicted LBD of the next learned clause is above a certain threshold. The MLB heuristic uses the LBDs of the last three learned clauses and their products as features to fit a linear function that predicts the LBD of the next learned clause. The performance of the MLR restart policy is better than Luby but worse than Glucose, as demonstrated on SATCOMP benchmarks.
	
	\subsubsection{Clause Deletion}
	Another entry point where learning can aid the CDCL solver is the clause deletion heuristic, which stands for the selection of useless clauses to be deleted learned from conflicts due to memory constraints. \cite{vaezipoorLearningClauseDeletion2020} formulates this task as a reinforcement learning problem and implements an OpenAI Gym compatible environment, \texttt{SAT-Gym}. Since the ultimate goal is to improve the running time of the SAT solver, the reward is related to the number of logical operations performed by the solver until an instance is solved.
	Similar to restarting, the clause deletion heuristic also relies on the LBD metric to evaluate clause quality.
	\cite{vaezipoorLearningClauseDeletion2020} optimizes a policy that outputs an LBD threshold as action by policy gradient so that all clauses with LBD values above the threshold are deleted.
	
	
	\subsection{Learning-aided SLS Solvers}
	\label{sec:SLS}
	Due to the straightforward framework of SLS solvers, the available heuristics for machine learning extension are fewer and simpler than CDCL solvers. For example, the variable selection does not need to predict the value of the variable, since the only operation is flipping. 
	
	\subsubsection{Variable Selection}
	\cite{yolcuLearningLocalSearch2019} proposed a variable selection heuristic for SLS solvers, which is computed by a graph neural network through deep reinforcement learning with a curriculum. The policy network, a GNN, takes as input a CNF formula in VCG form along with the current assignment and outputs a probability over variables, corresponding to their chances to be flipped in the next iteration. From the aspect of reinforcement learning, the reward is defined as whether the assignment satisfies the formula. The authors employed the REINFORCE algorithm to optimize the policy network. For faster convergence, they opted for curriculum learning and gradually increased the problem size. The empirical results of the learned heuristics are comparable to WalkSAT on small combinatorial instances, but it suffers considerable overhead since variable selection is required in every iteration.
	
	\subsubsection{Variable Initialization}
	A possible solution to circumvent the problem of high computational cost is using off-line training and focusing on less frequent operations, such as initialization, which only occurs after restarts.
	NLocalSAT~\cite{zhangNLocalSATBoostingLocal2020} takes this direction and boosts the performance of the SLS solver by guiding initialization assignments with a neural network. NLocalSAT feeds the CNF formula in LCG form into a GGCN for feature extraction, whose output is a predicted solution. The actual initialization process accepts the prediction for a high probability and preserves the ability for exploration. 
	Compared to~\cite{yolcuLearningLocalSearch2019}, the neural network is called only once for an instance. Within the timeout, NLocalSAT can solve more instances on multiple benchmarks, and this modification proves to be useful for various solvers.
	
	\section{Conclusion and Outlook}
	The integration of SAT solving and machine learning, as an emergent area of interest, has undergone rapid development. End-to-end SAT solvers have come to reality and been evaluated on random instances, even with better results on new instances of SAT competition. Meanwhile, recent years have also witnessed a line of works on the combination of learning-aided heuristics in existing solvers, yielding apparent improvements on search efficiency and effectiveness. 
	
	Nevertheless, several challenges remain to be solved, and the following problems merit further investigation. 
	Firstly, current standalone solvers can hardly scale to large instances, which is the common case in the real world. 
	Another barrier to using ML methods, especially neural networks, is their substantial computational time that may cancel the performance gain.
	Moreover, the intrinsic explainability problem of neural networks poses the question of how people should trust the predictions of an ML SAT solver when it does not provide precise proofs. Finally, we believe instance generation is a valuable yet underestimated topic, which could be improved by leveraging ML techniques that uncover the implicit structure of industrial instances and generate similar ones.
	
	
\end{spacing}

	\newpage
	\bibliographystyle{named}
	\bibliography{main}

\end{document}